\documentclass[runningheads]{llncs}

\usepackage{graphicx}
\usepackage{verbatim}
\usepackage{booktabs}
\usepackage{multirow}
\usepackage{xcolor}
\usepackage{makecell}

\usepackage[accsupp]{axessibility}  

\definecolor{gainGreen}{RGB}{0,120,60}   
\definecolor{gainRed}{RGB}{170,40,40}    
\definecolor{gainGray}{RGB}{120,120,120}

\usepackage{hyperref}

\usepackage{orcidlink}

\begin{document}

\title{Brain3D: EEG-to-3D Decoding of Visual Representations via Multimodal Reasoning}

\titlerunning{Brain3D}

\author{Emanuele Balloni\inst{1}\orcidlink{0000-0002-9510-5758} \and
Emanuele Frontoni\inst{2}\orcidlink{0000-0002-8893-9244} \and
Chiara Matti\inst{3} \and
Marina Paolanti\inst{2}\orcidlink{0000-0002-5523-7174} \and
Roberto Pierdicca\inst{1}\orcidlink{0000-0002-9160-834X} \and
Emiliano Santarnecchi\inst{4}\orcidlink{0000-0002-6533-7427}}

\authorrunning{E.~Balloni et al.}

\institute{Department of Civil Engineering Building and Architecture (DICEA), Università Politecnica delle Marche, Via Brecce Bianche 12, Ancona, 60131, Italy \and
Department of Political Sciences, Communication and International Relations, University of Macerata, Via Don Minzoni 22A, Macerata, 62100, Italy \and
Horizon Intelligence Labs \and
Harvard Medical School (HMS), Precision Neuromodulation Program \& Network Control Laboratory, Gordon Center for Medical Imaging, Department of Radiology, Massachusetts General Hospital, Horizon Intelligence Labs, Boston, MA, USA}

\maketitle

\sloppy

\begin{abstract}
Decoding visual information from electroencephalography (EEG) has recently achieved promising results, primarily focusing on reconstructing two-dimensional (2D) images from brain activity. However, the reconstruction of three-dimensional (3D) representations remains largely unexplored. This limits the geometric understanding and reduces the applicability of neural decoding in different contexts. To address this gap, we propose Brain3D, a multimodal architecture for EEG-to-3D reconstruction based on EEG-to-image decoding. It progressively transforms neural representations into the 3D domain using geometry-aware generative reasoning. Our pipeline first produces visually grounded images from EEG signals, then employs a multimodal large language model to extract structured 3D-aware descriptions, which guide a diffusion-based generation stage whose outputs are finally converted into coherent 3D meshes via a single-image-to-3D model. By decomposing the problem into structured stages, the proposed approach avoids direct EEG-to-3D mappings and enables scalable brain-driven 3D generation. We conduct a comprehensive evaluation comparing the reconstructed 3D outputs against the original visual stimuli, assessing both semantic alignment and geometric fidelity. Experimental results demonstrate strong performance of the proposed architecture, achieving up to 85.4\% 10-way Top-1 EEG decoding accuracy and 0.648 CLIPScore, supporting the feasibility of multimodal EEG-driven 3D reconstruction.

\keywords{Artificial Intelligence \and 3D \and Generative AI \and Neuroscience \and Computer Vision \and Computer Graphics \and EEG \and EEG-to-3D}
\end{abstract}

\section{Introduction}
\label{sec:intro}
The ability to reconstruct visual information from brain activity is a long-standing goal in both neuroscience and artificial intelligence (AI)~\cite{kneeland2025nsd,li2025brain,shah2025sam4em,schmors2025trace,wang2025zebra}. This has significant implications for brain-computer interfaces (BCIs), cognitive science, assistive technologies and human-AI interaction~\cite{nishimoto2011reconstructing,shen2019deep,kneeland2023reconstructing,bobrov2011brain}. The possibility of inferring what a person is perceiving or imagining directly from neural signals could facilitate more natural communication between humans and machines, and provide deeper insight into the mechanisms of visual cognition. 

Recent advances in deep learning and generative modeling have greatly enhanced the accuracy of neural decoding \cite{chen2023seeing,guo2025survey}, particularly in electroencephalography (EEG)-to-image reconstruction~\cite{ahmadieh2024visual,bai2024dreamdiffusion,lopez2025guess,takagi2023high}. These methods show that it is possible to extract meaningful semantic and perceptual information from non-invasive neural measurements. This is an important step towards creating useful brain-driven visual generation systems. However, human perception is inherently three-dimensional. Our interaction with the world relies on properties such as depth, geometry, spatial structure and viewpoint consistency, all of which extend beyond the scope of two-dimensional images~\cite{lu2025unfolding,welchman2016human}. Despite the fundamental three-dimensional nature of perception, most approaches to neural decoding are still limited to two-dimensional representation~\cite{fu2025brainvis,lopez2025guess,huo2024neuropictor,singh2024learning}. Although 2D reconstructions offer valuable semantic information, they lack explicit geometric structure and spatial consistency, which restricts their use in areas such as extended reality (XR), embodied AI, robotics and 3D simulation. Therefore, bridging neural activity and full 3D representation learning is a critical next step towards aligning computational models with the true structure of human visual cognition.
In this context, EEG is a highly relevant technique for addressing this challenge. As a non-invasive neuroimaging technique with temporal resolution in the millisecond range, EEG can capture the rapid neural processes involved in visual perception. Its portability, safety and affordability make it suitable for large-scale, real-world applications~\cite{bai2024dreamdiffusion,fu2025brainvis,cao2025eeg,singh2023eeg2image}. However, it is still very challenging to create meaningful 3D models from EEG signals due to three main factors: the indirect and noisy nature of EEG measurements, the ambiguity of geometric information in neural representations, and the absence of methodologies and benchmarks specifically designed to link EEG signals and 3D geometry.

Recent efforts have begun exploring EEG-to-3D reconstruction. In particular, Neuro-3D \cite{guo2025neuro} introduces a dedicated EEG-3D dataset and leverages diffusion priors to reconstruct colored 3D objects from brain signals. Mind2Matter \cite{deng2025mind2matter} proposes an end-to-end pipeline that maps EEG features into textual representations and then into 3D structures using layout-guided 3D Gaussians. In~\cite{masclef2025dual}, the authors aim to model depth perception through specialized neural streams designed to capture spatial reasoning from EEG. Xiang et al.~\cite{xiang2025electroencephalography} combines multi-task EEG representation learning with diffusion-guided Neural Radiance Field optimization to enforce stylistic consistency in the reconstructed 3D objects. However, these approaches require either specialized 3D stimuli, or attempt to learn direct, or end-to-end mappings from EEG to geometry. While promising, such strategies can be limited in terms of scalability, generalization, and often struggle to obtain geometrically consistent and perceptually detailed reconstructions from real-world EEG signals. We therefore propose that extending neural decoding to the 3D domain should be based on, rather than bypassing, the considerable progress made in EEG-to-image reconstruction.

In this study, we present Brain3D, a multimodal 3D decoding architecture that addresses the issue of EEG-to-3D reconstruction by redefining it as a progressive, cross-modal reasoning problem. Rather than mapping EEG signals directly to 3D geometry, Brain3D decomposes the task into three interconnected stages.
First, diffusion-guided image decoding is performed, transforming EEG signals into visually grounded image representations using state-of-the-art EEG-to-image models \cite{lopez2025guess,bai2024dreamdiffusion,fu2025brainvis,cao2025eeg}. This stage exploits the robustness and semantic richness of image-level neural decoding.
Then, a geometry-aware semantic reasoning module based on a Multimodal Large Language Model (MLLM) \cite{grattafiori2024llama} is introduced. Rather than considering the decoded images as the final output, they are projected into a structured semantic space. In this space, 3D-relevant attributes such as shape, spatial relationships and viewpoint cues are extracted through guided prompting and cross-attention reasoning.
Finally, we perform semantics-to-geometry generative modeling via a diffusion-based 3D synthesis stage. This structured semantic representation guides a generative process that produces, at first, 2D visual representations through diffusion \cite{patil2022stable}, to then create consistent 3D meshes \cite{xiang2025structured} while preserving the perceptual fidelity obtained from the neural signals. Through an iterative process of decomposition, Brain3D enables the progressive transfer of knowledge across modalities, bridging the gap between neural dynamics, visual perception, language reasoning and 3D geometry, without the need for direct EEG-to-mesh supervision. Furthermore, the proposed architecture is model-agnostic and can be integrated with various EEG-to-image backbones. We comprehensively evaluate Brain3D through quantitative and qualitative analysis, assessing geometric fidelity and semantic consistency with respect to the original visual stimuli. We also analyze the relationship between the quality of the intermediate EEG-to-image reconstruction and the performance of the final 3D output, providing insight into how visual encoding influences geometric synthesis, and perform an ablation study to assess the impact of the reconstruction steps.

Our main contributions can be summarized as follows:
\begin{itemize}
    \item We define the EEG-to-3D reconstruction process as a multimodal reasoning problem involving multiple stages, and we propose Brain3D: a structured 3D decoding architecture that establishes a connection between neural signals and geometry through progressive cross-modal alignment.
    \item We introduce a geometry-aware semantic reasoning module that uses MLLM-guided prompting and cross-modal attention to extract structured 3D attributes from EEG-derived images.
    \item We propose a semantics-to-geometry diffusion-based synthesis strategy that generates coherent, geometrically consistent 3D meshes based on structured semantic representations.
    \item We provide a comprehensive evaluation protocol that assesses geometric fidelity, semantic consistency and the impact of intermediate visual decoding quality on the performance of the final 3D reconstruction. This protocol demonstrates model-agnostic integration across multiple EEG-to-image backbones.
\end{itemize}

\section{Method}
\label{sec:materials}
\begin{figure}[!htb]
  \centering
  \includegraphics[width=\linewidth]{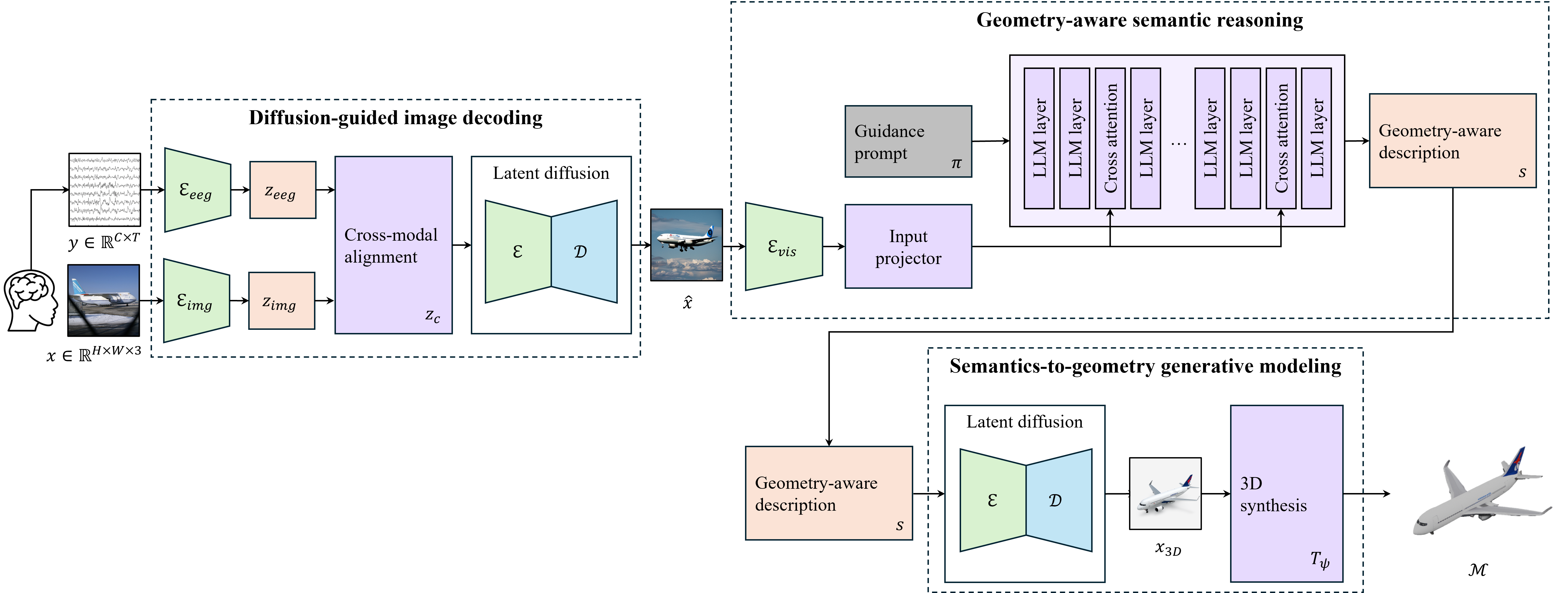}
  \caption{Overview of the proposed Brain3D architecture. Given an image and its corresponding EEG trial, the \textit{diffusion-guided image decoding} module first reconstructs a visually grounded image. Then, the \textit{geometry-aware semantic reasoning} stage then employs an MLLM to extract an object-centric, 3D-oriented textual description. Finally, the \textit{semantic-to-geometry generative modeling} module synthesizes a refined 2D image and lifts it into a 3D mesh representation.}
  \label{fig:brain3d_overview}
\end{figure}

Brain3D is a multimodal architecture for 3D mesh generation from EEG data, organized in three stages: EEG-to-image decoding, geometry-aware semantic reasoning and semantics-to-geometry generative modeling.
EEG signals are first decoded into an image representation, which is then transformed by a MLLM into a compact, 3D-oriented description. This description conditions a diffusion generator to produce a cleaner, more 3D-consistent image that serves as input to a single-image-to-3D model for mesh creation. This staged design promotes cross-modal transfer from EEG to vision, language, and geometry while avoiding an end-to-end EEG-to-3D mapping.
The overview is shown in Figure~\ref{fig:brain3d_overview}. 

In the following subsections, we detail the architecture steps, the evaluation procedure and metrics used in our experiments.

\subsection{Diffusion-guided image decoding}
The first stage of Brain3D transforms EEG activity into a visually grounded image representation that serves as the perceptual reference for subsequent multimodal reasoning. We introduce a unified diffusion-guided decoding module that factorizes the architectural constructs shared by recent state-of-the-art methods \cite{lopez2025guess,fu2025brainvis,bai2024dreamdiffusion,cao2025eeg} into three functional components: neural encoding, cross-modal alignment, and diffusion conditioning. Unlike prior task-specific EEG-to-image pipelines, we explicitly formalize their shared conditioning structure into a unified and modular interface suitable for downstream 3D reasoning.

\paragraph{Neural encoding.}
Let $\mathbf{y} \in \mathbb{R}^{C \times T}$ denote an EEG trial with $C$ channels and temporal length $T$, and let $\mathbf{x} \in \mathbb{R}^{H \times W \times 3}$ be the corresponding stimulus image available during training. Brain3D adopts the common two-stream paradigm of modern EEG-to-image systems, where neural and visual signals are embedded into a shared latent space.

The EEG signal is first processed by an encoder $\mathcal{E}_{\text{eeg}}$:
\begin{equation}
\mathbf{z}_{\text{eeg}} = \mathcal{E}_{\text{eeg}}(\mathbf{y}),
\end{equation}
which captures discriminative spatiotemporal patterns associated with the perceived stimulus. In parallel, the image is encoded through a visual encoder $\mathcal{E}_{\text{img}}$:
\begin{equation}
\mathbf{z}_{\text{img}} = \mathcal{E}_{\text{img}}(\mathbf{x}).
\end{equation}
This dual-encoding strategy provides a semantic reference during training, encouraging the neural embedding to align with the visual feature space. The proposed formulation is intentionally agnostic to the specific architectures used for $\mathcal{E}_{\text{eeg}}$ and $\mathcal{E}_{\text{img}}$, enabling compatibility with multiple EEG-to-image backbones.

\paragraph{Cross-modal alignment.}
To bridge neural and visual domains, we introduce an alignment operator $A(\cdot)$ that projects EEG features into the conditioning space of the generative prior:
\begin{equation}
\mathbf{z}_c = A(\mathbf{z}_{\text{eeg}}).
\end{equation}
During training, the alignment is guided by the visual embedding $\mathbf{z}_{\text{img}}$, encouraging semantic consistency between the two modalities via similarity-based supervision. Conceptually, the alignment module acts as a semantic normalization layer that harmonizes heterogeneous EEG encoders into a unified conditioning space, thereby stabilizing cross-modal generation. This approach enables plug-and-play integration within a shared generative architecture.

\paragraph{Diffusion conditioning.}
Given the aligned conditioning vector $\mathbf{z}_c$, image synthesis is performed through a conditional latent diffusion process. Starting from Gaussian noise $\mathbf{x}_T \sim \mathcal{N}(0, I)$, the denoising network $D_\theta$ reconstructs the image:
\begin{equation}
\hat{\mathbf{x}} = D_\theta(\mathbf{x}_T \mid \mathbf{z}_c).
\end{equation}
Classifier-free guidance is employed during sampling to balance perceptual realism and neural faithfulness. The diffusion prior injects strong natural image statistics, while the EEG-derived conditioning steers the reconstruction toward the semantic content encoded in the brain signals. The results is a semantically faithful reconstructed image $\hat{\mathbf{x}}$, that serves as the perceptual bridge for the subsequent geometry-aware semantic reasoning stage.

\subsection{Geometry-aware semantic reasoning}

The diffusion-guided decoding stage produces an image $\hat{\mathbf{x}}$ that preserves the semantic content inferred from EEG signals. However, directly feeding this image into a single-image-to-3D model often leads to unstable geometry due to missing structural cues, artifacts and distortions. To address this limitations, the geometry-aware semantic reasoning module converts the decoded image into a structured, 3D-oriented textual representation suitable for downstream generative modeling.

\paragraph{MLLM-based visual reasoning.}
Given the reconstructed image $\hat{\mathbf{x}}$, we employ a MLLM (specifically LLaMA~3.2 Vision 90B) to extract a detailed object-centric description. The MLLM is treated as a conditional function
\begin{equation}
\mathbf{s} = F_{\text{MLLM}}(\hat{\mathbf{x}}, \pi),
\end{equation}
where $\pi$ denotes a task-specific prompting template and $\mathbf{s}$ is the resulting semantic description. Unlike generic captioning, the goal is not linguistic completeness but 3D-relevant semantic structuring. The MLLM leverages cross-attention between visual tokens and language priors to infer object attributes such as shape, material, and geometric cues that are beneficial for 3D reconstruction.

\paragraph{Prompt conditioning strategy.}
A key design choice of Brain3D is the use of a strongly constrained prompting protocol that explicitly biases the MLLM toward geometry-preserving descriptions. The template $\pi$ is defined as:

\begin{quote}\small
\textbf{System prompt:} \textit{You are an expert in generating prompts for text-to-2D diffusion models.}

\textbf{User prompt:} \textit{Create a prompt to be fed to the text-to-image model. The prompt should describe only the single main object in the image in high details. Focus on every aspect of the main object, such as the shape, color, material, and style. The prompt should be long. The prompt should describe the main object as a 3D model. Do not describe anything else other than the main object. The object needs to be the only element in the prompt. Force a white background. Do not use bullet points. Return only the prompt text. No introduction, explanations or formatting.}
\end{quote}

This constrained instruction serves two purposes: (i) it suppresses background and contextual noise that may hinder geometric reconstruction, and (ii) it encourages the emergence of explicit 3D-aware descriptors.

\paragraph{Structured semantic projection.}
The resulting text $\mathbf{s}$ can be interpreted as a structured semantic projection of the EEG-derived visual content into language space. Formally, this module implements the mapping
\begin{equation}
\mathbf{s} = \mathcal{R}(\mathbf{y}) = F_{\text{MLLM}}\!\left(D_\theta(\mathbf{x}_T \mid A(E_{\text{eeg}}(\mathbf{y}))), \pi\right),
\end{equation}
which bridges neural activity, visual evidence, and language reasoning within a unified pipeline.


The output of this module is a high-fidelity textual prompt $\mathbf{s}$ describing the main object as a clean, isolated 3D entity, which is used to condition the final generative stage.

\subsection{Semantic-to-geometry generative modeling}
The geometry-aware semantic reasoning module produces a structured textual prompt $\mathbf{s}$ that explicitly describes the main object with 3D-relevant attributes. The final stage of Brain3D leverages this representation to progressively lift semantics into geometry. Instead of directly predicting 3D from language, we adopt a two-step generative strategy that first synthesizes a clean 2D visual proxy and subsequently reconstructs the corresponding 3D shape. This decomposition improves stability and allows the pipeline to benefit from mature diffusion priors.

\paragraph{Text-to-image diffusion synthesis.}
Given the semantic description $\mathbf{s}$, we generate a refined object-centric image (using Stable Diffusion~3.5 Medium). The diffusion model is treated as a conditional generator
\begin{equation}
x_\text{3D} = G_{\phi}(\mathbf{s}),
\end{equation}
where $G_{\phi}$ denotes the pretrained text-to-image diffusion backbone and $x_\text{3D}$ is the synthesized image. 
Thanks to the constrained prompt produced in the previous stage (object-only, white background, 3D-focused description), the generated image exhibits reduced background clutter, improved object completeness, and stronger geometric consistency compared to the raw EEG reconstruction. This step can be interpreted as a semantic refinement and normalization process that prepares the visual input for reliable 3D lifting.

\paragraph{Single-image-to-3D reconstruction.}
The refined image $x_\text{3D}$ is then converted into a 3D representation (using TRELLIS), a feed-forward single-image-to-3D model. We model this stage as
\begin{equation}
\mathcal{M} = T_{\psi}(x_\text{3D}),
\end{equation}
where $T_{\psi}$ denotes the 3D generation network and $\mathcal{M}$ is the reconstructed 3D mesh. The network infers volumetric and surface structure from the monocular image by leveraging learned 3D priors, enabling the recovery of coherent geometry from a single view.
The success of this stage depends critically on the object-centric quality of $x_\text{3D}$. By enforcing semantic purification in the previous modules and diffusion-based refinement here, Brain3D provides the 3D generation network with inputs that better satisfy the assumptions of single-image 3D reconstruction, leading to more stable meshes and improved geometric fidelity.

\subsection{Evaluation protocol}
\label{subsec:eval}
We evaluate Brain3D from both semantic and geometric perspectives by comparing the reconstructed outputs against the original visual stimuli. Given the absence of direct EEG-to-3D ground-truth supervision, we adopt a rendering-based protocol that enables consistent comparison in the image domain while still reflecting 3D structural quality.

For each reconstructed mesh $\mathcal{M}$, we generate a set of canonical views using Blender. Specifically, six viewpoints are rendered to cover the full object extent: \emph{front}, \emph{front-left}, \emph{left}, \emph{back}, \emph{right}, and \emph{front-right}. The cameras are placed on a circular trajectory around the object with a fixed elevation and an azimuth step of $30^\circ$, ensuring uniform coverage of the visible geometry.

Formally, let
\[
\mathcal{V}(\mathcal{M}) = \{\mathbf{v}_1,\dots,\mathbf{v}_6\}
\]
denote the set of rendered views. Each view is centered on the object and generated under consistent lighting and background conditions to minimize rendering bias. Each rendered view is compared against the corresponding ground-truth stimulus image $\mathbf{x}$ from the EEGCVPR40 dataset (detailed in Sec. \ref{subsec:exp_setup}). Although only a single reference image is available, the multi-view evaluation allows us to assess whether the reconstructed geometry maintains semantic consistency across viewpoints. This protocol follows common practice in single-image-to-3D evaluation \cite{xiang2025structured,go2025vist3a,poole2022dreamfusion,liu2024one,xu2024grm}.

\paragraph{Evaluation metrics.}

We compute both semantic-consistency and perceptual-quality measures on the rendered views and aggregate them across viewpoints and samples. For a given object, view-level metrics are first computed between $\mathbf{x}$ and each rendered view $\mathbf{v}_i$, then averaged across $i=1,\dots,6$ to obtain a per-object score. The leveraged metrics are: CLIPScore, Learned Perceptual Image Patch Similarity (LPIPS), Inception Score (IS), Fr\'echet Inception Distance (FID) and Top-$k$ $n$-way accuracy.

CLIPScore measures semantic alignment between $\mathbf{x}$ and each view $\mathbf{v}_i$ using CLIP image embeddings. Specifically, we use CLIP ViT-B/16 to extract normalized features $f(\cdot)$ and compute cosine similarity:
\begin{equation}
\text{CLIPScore}(\mathbf{x},\mathbf{v}_i)=
\frac{f(\mathbf{x})^\top f(\mathbf{v}_i)}{\|f(\mathbf{x})\|_2 \, \|f(\mathbf{v}_i)\|_2}.
\end{equation}
We use LPIPS with the AlexNet backbone and, for each view, we compute
\begin{equation}
\text{LPIPS}(\mathbf{x},\mathbf{v}_i),
\end{equation}
and report per-object averages, alongside global mean and standard deviation over the test set.
Furthermore, we compute IS and FID over the set of all rendered views.

To quantify semantic decoding consistency under the standard retrieval setting, we also evaluate Top-$k$ $n$-way accuracy using a classical classifier-based protocol. We use an ImageNet-pretrained classifier to obtain class-probability vectors for each rendered view and for the ground-truth image. Let $\mathbf{p}_i \in \mathbb{R}^{K}$ be the predicted class distribution for view $\mathbf{v}_i$ and let $c^\star$ be the predicted class index for $\mathbf{x}$ (obtained from the same classifier). For each trial, we sample an $n$-way candidate set consisting of the positive class $c^\star$ and $n-1$ negatives, and count the reconstruction as correct if the positive class is ranked within the top-$k$ scores. We repeat this procedure for multiple random negative samplings (i.e., multiple trials) and report the mean and standard deviation across trials, then average across the six views.


\section{Experiments and results}
\label{sec:exp}

\subsection{Experimental setup}
\label{subsec:exp_setup}
\paragraph{Dataset}
We leveraged the EEGCVPR40 dataset for our experiments~\cite{spampinato2017deep,palazzo2020decoding}. It is among the most widely used benchmarks for EEG-to-image reconstruction \cite{singh2024learning,lopez2025guess,bai2024dreamdiffusion}. The dataset provides EEG recordings from 6 participants viewing 2,000 images drawn from 40 ImageNet object categories \cite{deng2009imagenet}. Each category includes 50 images, presented sequentially for 0.5 seconds per trial at 1 kHz. After each block of 50 images, a 10-second pause was inserted to reduce fatigue and allow for reset. EEG activity was acquired with a 128-channel system (ActiCAP 128ch). The stimulus set spans a broad range of visual categories, including animals (e.g., dogs, cats, elephants), vehicles (e.g., airliners, bicycles, cars), and everyday objects (e.g., computers, chairs, mugs), ensuring broad semantic coverage. We follow the original split protocol \cite{spampinato2017deep}, with training, validation, and test sets corresponding to 80\%, 10\%, and 10\% of the data, respectively.


\paragraph{Implementation details}
All experiments were conducted on a workstation equipped with an NVIDIA H100 GPU with 80\,GB of VRAM, an Intel Xeon Platinum 8481C CPU, and 240\,GB of RAM. The operating system was Ubuntu 22.04. The software environment was based on Python~3.12 with PyTorch~2.7.1 and CUDA~12.9. 

For the Diffusion-guided image decoding module, we integrated four state-of-the-art EEG-to-image reconstruction frameworks: Guess What I Think (GWIT) \cite{lopez2025guess}, BrainVis \cite{fu2025brainvis}, DreamDiffusion \cite{bai2024dreamdiffusion}, and EEG-CLIP \cite{cao2025eeg}. To ensure a fair comparison and preserve the characteristics of each backbone, all hyperparameters and architectural configurations were kept identical to their original implementations. The Geometry-aware semantic reasoning module employs the LLaMA~3.2 Vision 90B MLLM to generate object-centric textual descriptions from the reconstructed images. In the Semantic-to-geometry generative modeling stage, Stable Diffusion~3.5 Medium is used to synthesize a refined 2D image from the generated textual prompt with, 30 inference steps and a CFG of 4.5. The resulting image is then converted into a 3D representation using Microsoft TRELLIS, with a 1024 texture resolution. All remaining parameters follow the default settings of the respective original implementations.

\subsection{Results}
We evaluate Brain3D using the protocol described in Sec.~\ref{subsec:eval}, comparing the rendered views of the reconstructed 3D objects with the original ground-truth stimulus images. In addition, we analyze the influence of the intermediate EEG-to-image decoding stage by comparing the rendered views against the images produced by the EEG-to-image models themselves.

\begin{table*}[!htb]
\centering
\caption{Quantitative evaluation of Brain3D. Metrics are computed between the six rendered views of the reconstructed 3D mesh and the ground-truth stimulus image.}
\label{tab:results_gt}
\resizebox{\textwidth}{!}{%
\setlength{\tabcolsep}{6pt}
\begin{tabular}{@{}lccccccccc@{}}
\toprule
\multirow{2}{*}{\textbf{Backbone}} 
& \multicolumn{1}{c}{\textbf{2-way Acc $\uparrow$}} 
& \multicolumn{2}{c}{\textbf{10-way Acc $\uparrow$}} 
& \multicolumn{2}{c}{\textbf{50-way Acc $\uparrow$}} 
& \multirow{2}{*}{\textbf{CLIPScore $\uparrow$}} 
& \multirow{2}{*}{\textbf{IS $\uparrow$}} 
& \multirow{2}{*}{\textbf{FID $\downarrow$}} 
& \multirow{2}{*}{\textbf{LPIPS $\downarrow$}} \\
\cmidrule(lr){2-2}
\cmidrule(lr){3-4}
\cmidrule(lr){5-6}
& \textbf{Top-1} 
& \textbf{Top-1} & \textbf{Top-2} 
& \textbf{Top-1} & \textbf{Top-2} 
& & & & \\
\midrule
BrainVis & 0.880 & 0.706 & 0.796 & 0.578 & 0.649 & 0.617 & 16.590 & 204.015 & 0.789 \\
DreamDiffusion & 0.730 & 0.314 & 0.480 & 0.164 & 0.206 & 0.564 & 14.871 & 232.256 & \textbf{0.780} \\
EEG-CLIP & 0.857 & 0.655 & 0.742 & 0.545 & 0.602 & 0.608 & 17.173 & 156.631 & 0.788 \\
GWIT & \textbf{0.946} & \textbf{0.854} & \textbf{0.906} & \textbf{0.763} & \textbf{0.822} & \textbf{0.648} & \textbf{17.195} & \textbf{153.295} & 0.783 \\
\bottomrule
\end{tabular}}
\end{table*}


Table~\ref{tab:results_gt} reports the results obtained when comparing the rendered views of the reconstructed meshes against the original stimulus images. Among the evaluated EEG-to-image backbones, the configuration based on \textit{GWIT} consistently achieves the strongest performance across most semantic metrics. In particular, it obtains the highest 10-way Top-1 accuracy (0.8539), the highest CLIPScore (0.6478), and the lowest FID (153.30), indicating that the reconstructed 3D objects maintain strong semantic alignment with the original stimuli while also producing more realistic visual distributions. Similar trends are observed for more challenging retrieval settings, where GWIT achieves 50-way Top-1 accuracy of 0.7633 and 50-way Top-2 accuracy of 0.8224.
The \emph{BrainVis} and \emph{EEG-CLIP} configurations also demonstrate competitive performance, with CLIPScores of 0.6172 and 0.6083 respectively, suggesting that the semantic information extracted from the EEG signals is largely preserved throughout the generative pipeline. In contrast, the \emph{DreamDiffusion} backbone produces significantly lower decoding accuracy (10-way Top-1 of 0.3141) and the highest FID (232.26), indicating that weaker intermediate image reconstructions propagate to the final 3D generation stage.
Perceptual quality metrics show a similar trend. IS ranges between 14.87 and 17.20 across models, with the highest values obtained for GWIT and EEG-CLIP. LPIPS values remain relatively consistent across all configurations, suggesting comparable perceptual similarity levels between reconstructed views and the ground-truth images.


These results are consistent with the metrics reported in the original EEG-to-image reconstruction studies. In particular, the \emph{GWIT} backbone, which achieves the strongest performance in its original setting, also leads to the best results in the proposed Brain3D architecture, while \emph{DreamDiffusion}, which exhibits comparatively lower decoding accuracy, produces weaker downstream performance. This alignment suggests that the quality of the intermediate EEG-to-image decoding stage largely determines the final reconstruction fidelity. Nevertheless, despite these differences in upstream performance, Brain3D is able to generate coherent 3D representations across all configurations, demonstrating its ability to propagate semantic information from EEG signals to structured 3D geometry even under challenging reconstruction scenarios.

\begin{figure*}[!htb]
\centering
\includegraphics[width=\textwidth]{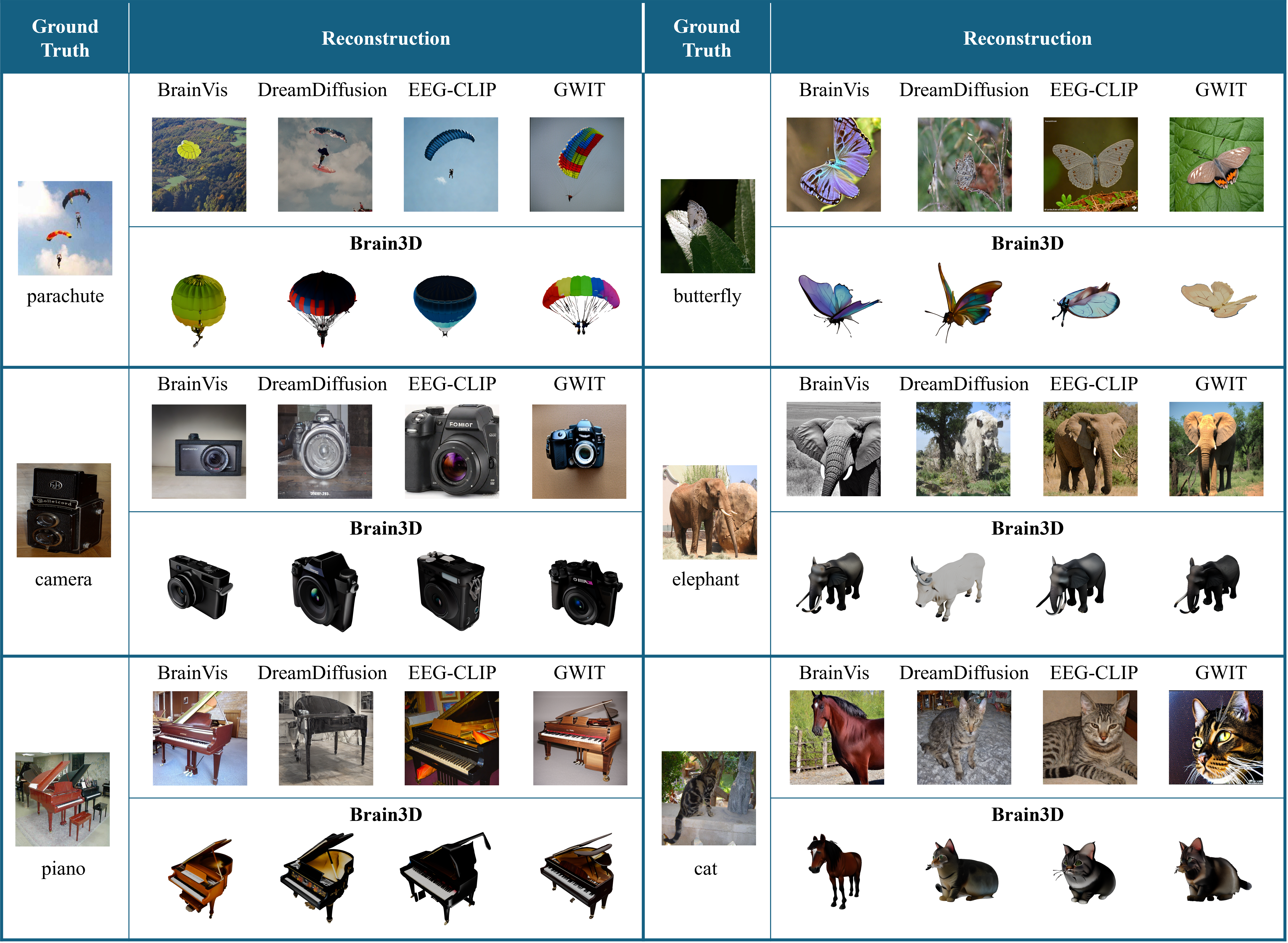}
\caption{
Qualitative examples of Brain3D reconstructions across multiple object categories. 
For each example, the ground-truth stimulus is shown on the left, followed by EEG-to-image reconstructions produced by different decoding models, and the corresponding 3D objects generated by Brain3D.
}
\label{fig:qualitative_results}
\end{figure*}

Figure~\ref{fig:qualitative_results} shows qualitative examples of the reconstructed 3D objects across multiple categories. In most cases, the generated 3D models capture the main semantic characteristics of the target objects, preserving their overall shape and recognizable structural components. For instance, parachutes are reconstructed with their characteristic canopy structure, cameras maintain their compact body and lens configuration, and animals such as elephants and cats exhibit distinctive body proportions and anatomical features.
Some failure cases are also reported, such as the \emph{DreamDiffusion} elephant and the \emph{BrainVis} cat, where the reconstructed 3D model does not correspond to the correct object category. These failures originate from the initial Diffusion-guided image decoding stage, where the intermediate image reconstruction does not correctly predict the target class. This behavior is expected, as earlier EEG-to-image approaches exhibit lower decoding accuracy compared to more recent methods. Nevertheless, when the intermediate reconstruction preserves the correct semantic information, Brain3D is able to consistently generate coherent 3D representations that reflect the structure of the original stimuli.


\begin{table*}[t]
\centering
\caption{Impact of the EEG-to-image decoder on the final 3D reconstruction. Metrics are computed between the rendered views of the reconstructed meshes and the images generated by the Diffusion-guided image decoding module. Gains with respect to the ground-truth evaluation (Table~\ref{tab:results_gt}) are also reported.}
\label{tab:results_rec}
\resizebox{\textwidth}{!}{%
\setlength{\tabcolsep}{6pt}
\begin{tabular}{@{}lccccccccc@{}}
\toprule
\multirow{2}{*}{\textbf{Backbone}} 
& \multicolumn{1}{c}{\textbf{2-way Acc $\uparrow$}} 
& \multicolumn{2}{c}{\textbf{10-way Acc $\uparrow$}} 
& \multicolumn{2}{c}{\textbf{50-way Acc $\uparrow$}} 
& \multirow{2}{*}{\textbf{CLIPScore $\uparrow$}} 
& \multirow{2}{*}{\textbf{IS $\uparrow$}} 
& \multirow{2}{*}{\textbf{FID $\downarrow$}} 
& \multirow{2}{*}{\textbf{LPIPS $\downarrow$}} \\
\cmidrule(lr){2-2}
\cmidrule(lr){3-4}
\cmidrule(lr){5-6}
& \textbf{Top-1} 
& \textbf{Top-1} & \textbf{Top-2} 
& \textbf{Top-1} & \textbf{Top-2} 
& & & & \\
\midrule
\noalign{\vskip 3pt}
BrainVis
& \makecell{0.956 \\ {\color{gainGreen}\scriptsize +0.076}}
& \makecell{0.860 \\ {\color{gainGreen}\scriptsize +0.154}}
& \makecell{0.920 \\ {\color{gainGreen}\scriptsize +0.124}}
& \makecell{0.759 \\ {\color{gainGreen}\scriptsize +0.181}}
& \makecell{0.825 \\ {\color{gainGreen}\scriptsize +0.176}}
& \makecell{0.708 \\ {\color{gainGreen}\scriptsize +0.091}}
& \makecell{16.590 \\ \color{gainGray}\scriptsize 0.000}
& \makecell{187.696 \\ {\color{gainGreen}\scriptsize -16.319}}
& \makecell{0.806 \\ {\color{gainRed}\scriptsize +0.017}} \\
\noalign{\vskip 5pt}
\hline
\noalign{\vskip 5pt}
DreamDiffusion
& \makecell{0.862 \\ {\color{gainGreen}\scriptsize +0.132}}
& \makecell{0.604 \\ {\color{gainGreen}\scriptsize +0.290}}
& \makecell{0.734 \\ {\color{gainGreen}\scriptsize +0.254}}
& \makecell{0.426 \\ {\color{gainGreen}\scriptsize +0.262}}
& \makecell{0.517 \\ {\color{gainGreen}\scriptsize +0.311}}
& \makecell{0.636 \\ {\color{gainGreen}\scriptsize +0.072}}
& \makecell{14.871 \\ \color{gainGray}\scriptsize 0.000}
& \makecell{223.501 \\ {\color{gainGreen}\scriptsize -8.755}}
& \makecell{0.791 \\ {\color{gainRed}\scriptsize +0.011}} \\
\noalign{\vskip 5pt}
\hline
\noalign{\vskip 5pt}
EEG-CLIP
& \makecell{0.980 \\ {\color{gainGreen}\scriptsize +0.123}}
& \makecell{0.920 \\ {\color{gainGreen}\scriptsize +0.265}}
& \makecell{0.961 \\ {\color{gainGreen}\scriptsize +0.219}}
& \makecell{0.841 \\ {\color{gainGreen}\scriptsize +0.296}}
& \makecell{0.899 \\ {\color{gainGreen}\scriptsize +0.297}}
& \makecell{0.712 \\ {\color{gainGreen}\scriptsize +0.104}}
& \makecell{17.173 \\ \color{gainGray}\scriptsize 0.000}
& \makecell{170.292 \\ {\color{gainRed}\scriptsize +13.661}}
& \makecell{0.789 \\ {\color{gainGreen}\scriptsize +0.001}} \\
\noalign{\vskip 5pt}
\hline
\noalign{\vskip 5pt}
GWIT
& \makecell{0.977 \\ {\color{gainGreen}\scriptsize +0.031}}
& \makecell{0.905 \\ {\color{gainGreen}\scriptsize +0.051}}
& \makecell{0.953 \\ {\color{gainGreen}\scriptsize +0.047}}
& \makecell{0.815 \\ {\color{gainGreen}\scriptsize +0.052}}
& \makecell{0.876 \\ {\color{gainGreen}\scriptsize +0.054}}
& \makecell{0.736 \\ {\color{gainGreen}\scriptsize +0.088}}
& \makecell{17.437 \\ {\color{gainGreen}\scriptsize +0.242}}
& \makecell{101.031 \\ {\color{gainGreen}\scriptsize -52.264}}
& \makecell{0.805 \\ {\color{gainRed}\scriptsize +0.022}} \\

\bottomrule
\end{tabular}}
\end{table*}

\paragraph{Impact of the EEG-to-image intermediate representations}
To better understand how the how the intermediate visual decoding stage affects the final 3D reconstruction, we also compared the views rendered from the generated 3D models with the images produced directly by the Diffusion-guided image decoding module. The results, reported in Table~\ref{tab:results_rec}, consistently show higher semantic alignment across all backbones compared to the ground-truth evaluation. Substantial improvements can be seen across all N-way Top-k accuracy metrics. For example, EEG-CLIP exhibits gains of +0.265 and +0.296 for the 10-way and 50-way Top-1 accuracies respectively, while DreamDiffusion shows even larger improvements (+0.290 and +0.262), reflecting the recovery of semantic consistency once the intermediate image representation is used as reference. Similarly, GWIT achieves the strongest generative fidelity, reaching the lowest FID (101.03) with a large improvement of $-52.26$ compared to the ground-truth evaluation.
This behavior indicates that the 3D generation stage preserves most of the semantic structure present in the intermediate reconstructed images. In other words, once the EEG signal has been translated into a coherent visual representation, the subsequent semantic reasoning and generative stages introduce only limited degradation. The gap observed between the two evaluation settings therefore primarily reflects the difficulty of the EEG-to-image decoding task rather than the performance of the later stages of the architecture.

These results highlight two important observations. Firstly, the proposed Brain3D architecture effectively converts semantic information from neural signals into coherent 3D structures, achieving strong alignment with the original visual stimuli. Secondly, the quality of the intermediate EEG-to-image reconstruction remains the main bottleneck of the architecture, with stronger visual decoders leading to consistently improved 3D reconstruction performance. To address this issue, the Brain3D architecture has been designed to be model-agnostic, decoupling the diffusion-guided image decoding stage from the subsequent reasoning and generation modules. This design choice enables the architecture to accommodate the current limitations of EEG-to-image models and to adapt easily to future advances in neural decoding methods.


\subsection{Ablation study}
To assess the relevance of the \emph{Geometry-aware semantic reasoning} and \emph{Semantic-to-geometry generative modeling} modules we performed an ablation study by removing both stages and generating 3D models directly from the images produced by the Diffusion-guided image decoding module. We evaluate this setting with the same protocol and metrics used for the full pipeline. Table~\ref{tab:ablation_full_vs_direct} reports the full Brain3D scores (top line) together with the gain/loss with respect to the ablated configuration (second line).

\begin{table*}[t]
\centering
\caption{Ablation study evaluating the effectiveness of the full Brain3D architecture against the configuration with 3D models generated directly from the EEG-to-image reconstructions. For each backbone, we report the results obtained with the full Brain3D architecture, the ablated configuration, and the gain/loss obtained by the full pipeline.}
\label{tab:ablation}
\resizebox{\textwidth}{!}{%
\setlength{\tabcolsep}{6pt}
\begin{tabular}{@{}llccccccccc@{}}
\toprule
\multirow{2}{*}{\textbf{Backbone}} & \multirow{2}{*}{\textbf{Setting}}
& \multicolumn{1}{c}{\textbf{2-way Acc $\uparrow$}}
& \multicolumn{2}{c}{\textbf{10-way Acc $\uparrow$}}
& \multicolumn{2}{c}{\textbf{50-way Acc $\uparrow$}}
& \multirow{2}{*}{\textbf{CLIPScore $\uparrow$}}
& \multirow{2}{*}{\textbf{IS $\uparrow$}}
& \multirow{2}{*}{\textbf{FID $\downarrow$}}
& \multirow{2}{*}{\textbf{LPIPS $\downarrow$}} \\
\cmidrule(lr){3-3}
\cmidrule(lr){4-5}
\cmidrule(lr){6-7}
& 
& \textbf{Top-1}
& \textbf{Top-1} & \textbf{Top-2}
& \textbf{Top-1} & \textbf{Top-2}
& & & & \\
\midrule

\multirow{3}{*}{BrainVis}
& Full architecture 
& \textbf{0.880} & \textbf{0.706} & \textbf{0.796} & \textbf{0.578} & \textbf{0.649} & \textbf{0.617} & \textbf{16.590} & \textbf{204.015} & \textbf{0.789} \\
& Direct image-to-3D 
& 0.866 & 0.667 & 0.760 & 0.538 & 0.607 & 0.609 & 12.724 & 228.799 & 0.805 \\
& Gain/loss
& {\color{gainGreen}+0.014}
& {\color{gainGreen}+0.039}
& {\color{gainGreen}+0.036}
& {\color{gainGreen}+0.040}
& {\color{gainGreen}+0.042}
& {\color{gainGreen}+0.008}
& {\color{gainGreen}+3.866}
& {\color{gainGreen}-24.784}
& {\color{gainGreen}-0.016} \\

\addlinespace[4pt]

\multirow{3}{*}{DreamDiffusion}
& Full architecture
& \textbf{0.730} & \textbf{0.314} & \textbf{0.480} & \textbf{0.164} & 0.206 & \textbf{0.564} & \textbf{14.871} & \textbf{232.256} & \textbf{0.780} \\
& Direct image-to-3D
& 0.692 & 0.305 & 0.451 & 0.156 & \textbf{0.212} & 0.563 & 8.864 & 280.717 & 0.808 \\
& Gain/loss
& {\color{gainGreen}+0.038}
& {\color{gainGreen}+0.009}
& {\color{gainGreen}+0.029}
& {\color{gainGreen}+0.008}
& {\color{gainRed}-0.006}
& {\color{gainGreen}+0.001}
& {\color{gainGreen}+6.007}
& {\color{gainGreen}-48.461}
& {\color{gainGreen}-0.028} \\

\addlinespace[4pt]

\multirow{3}{*}{EEG-CLIP}
& Full architecture
& \textbf{0.857} & \textbf{0.655} & \textbf{0.742} & 0.545 & \textbf{0.602} & \textbf{0.608} & 17.173 & \textbf{156.631} & \textbf{0.788} \\
& Direct image-to-3D
& 0.851 & 0.645 & 0.723 & \textbf{0.552} & 0.599 & 0.606 & \textbf{17.303} & 165.348 & 0.792 \\
& Gain/loss
& {\color{gainGreen}+0.006}
& {\color{gainGreen}+0.010}
& {\color{gainGreen}+0.019}
& {\color{gainRed}-0.007}
& {\color{gainGreen}+0.003}
& {\color{gainGreen}+0.002}
& {\color{gainRed}-0.130}
& {\color{gainGreen}-8.717}
& {\color{gainGreen}-0.004} \\

\addlinespace[4pt]

\multirow{3}{*}{GWIT}
& Full architecture
& \textbf{0.946} & \textbf{0.854} & \textbf{0.906} & \textbf{0.763} & \textbf{0.822} & \textbf{0.648} & \textbf{17.195} & \textbf{153.295} & \textbf{0.783} \\
& Direct image-to-3D
& \textbf{0.946} & 0.836 & 0.893 & 0.733 & 0.801 & 0.627 & 14.567 & 183.566 & 0.808 \\
& Gain/loss
& {\color{gainGray}0.000}
& {\color{gainGreen}+0.018}
& {\color{gainGreen}+0.013}
& {\color{gainGreen}+0.030}
& {\color{gainGreen}+0.021}
& {\color{gainGreen}+0.021}
& {\color{gainGreen}+2.628}
& {\color{gainGreen}-30.271}
& {\color{gainGreen}-0.025} \\

\bottomrule
\end{tabular}}
\label{tab:ablation_full_vs_direct}
\end{table*}

The results show that the complete Brain3D pipeline consistently improves both semantic alignment and generative fidelity compared to directly generating 3D models from the EEG-reconstructed images. In particular, the full architecture yields higher N-way Top-k accuracies for most backbones, indicating stronger semantic consistency between the reconstructed 3D objects and the original stimuli. For example, the BrainVis backbone improves the 10-way Top-1 accuracy from $0.667$ to $0.706$ and the 50-way Top-1 accuracy from $0.538$ to $0.578$. Similar improvements are observed for GWIT, where the 50-way Top-1 accuracy increases from $0.733$ to $0.763$.
The benefits of the full architecture are even more apparent when it comes to generative quality metrics. In particular, the FID scores are significantly reduced across all backbones when the reasoning and generative modules are included. For instance, the GWIT configuration improves from $183.57$ to $153.30$, while DreamDiffusion shows an even larger reduction from $280.72$ to $232.26$. Similar improvements are observed for LPIPS, indicating that the reconstructed views become perceptually closer to the ground-truth stimuli.
These values highlight that the intermediate semantic reasoning stage effectively extracts structured object descriptions from the intermediate reconstructions, enabling the subsequent generative process to better preserve object-level semantics, and helps mitigate the noise and ambiguities present in the EEG-to-image reconstructions before the final 3D generation step.

Taken together, these results confirm that directly converting EEG-reconstructed images into 3D geometry is insufficient to achieve optimal reconstruction quality. Instead, the \textit{Geometry-aware semantic reasoning} and \textit{Semantic-to-geometry generative modeling} modules play a crucial role in refining the intermediate representations and guiding the generative process. This highlights the importance of the full Brain3D architecture for reliably translating neural signals into coherent and semantically consistent 3D object representations.

\section{Conclusions and future works}
\label{sec:conclusions}
In this work, we introduced Brain3D, a multimodal architecture for reconstructing 3D object representations directly from EEG signals. Rather than learning a direct mapping from neural activity to geometry, the proposed architecture decomposes the EEG-to-3D problem into a sequence of structured cross-modal transformations.
Experimental results showed that the architecture can successfully preserve object-level semantics from neural signals to the final 3D outputs. Across different EEG-to-image backbones, the reconstructed meshes maintain strong alignment with the original visual stimuli, while quantitative evaluations confirm consistent performance across multiple retrieval and perceptual metrics. The analysis of intermediate representations further indicated that the fidelity of the EEG-to-image stage remains the main factor influencing the quality of the final reconstructions. Furthermore, the ablation study performed highlighted the contribution of the Geometry-aware semantic reasoning and Semantic-to-geometry generative modeling modules. When these stages are removed and 3D shapes are generated directly from EEG-reconstructed images, both semantic retrieval accuracy and generative quality degrade.

These findings highlight an important insight: while EEG-to-image decoding remains the primary bottleneck of the overall system, the proposed multimodal reasoning pipeline effectively stabilizes the subsequent generation stages and preserves the semantic information extracted from neural signals. For this reason, Brain3D was intentionally designed as a model-agnostic architecture that decouples neural decoding from the downstream reasoning and generative modules. This design enables the architecture to operate under the limitations of current EEG-based visual decoding methods while remaining fully compatible with future improvements in neural representation learning.

Several directions for future research emerge from this work. First, extending Brain3D to reconstruct more complex scenes containing multiple objects and spatial relationships would bring neural decoding closer to real-world visual perception. Second, incorporating temporal neural dynamics could enable the reconstruction of dynamic 3D content from continuous brain activity. Third, integrating more advanced multimodal foundation models may further improve the extraction of geometry-aware semantic representations. Finally, future work may explore higher-resolution 3D generation models and alternative neural decoding techniques to enhance geometric fidelity and enable more detailed brain-driven 3D reconstructions.

%
%
\bibliographystyle{splncs04}
\bibliography{main}

@String(NeurIPS = {Adv. Neural Inform. Process. Syst.})

@String(ICASSP=	{ICASSP})

@String(NeurIPS = {NeurIPS})

@article{nishimoto2011reconstructing,
  title={Reconstructing visual experiences from brain activity evoked by natural movies},
  author={Nishimoto, Shinji and Vu, An T and Naselaris, Thomas and Benjamini, Yuval and Yu, Bin and Gallant, Jack L},
  journal={Current biology},
  volume={21},
  number={19},
  pages={1641--1646},
  year={2011},
  publisher={Elsevier}
}

@article{shen2019deep,
  title={Deep image reconstruction from human brain activity},
  author={Shen, Guohua and Horikawa, Tomoyasu and Majima, Kei and Kamitani, Yukiyasu},
  journal={PLoS computational biology},
  volume={15},
  number={1},
  pages={e1006633},
  year={2019},
  publisher={Public Library of Science San Francisco, CA USA}
}

@article{kneeland2023reconstructing,
  title={Reconstructing seen images from human brain activity via guided stochastic search},
  author={Kneeland, Reese and Ojeda, Jordyn and St-Yves, Ghislain and Naselaris, Thomas},
  journal={ArXiv},
  pages={arXiv--2305},
  year={2023}
}

@article{ahmadieh2024visual,
  title={Visual image reconstruction based on EEG signals using a generative adversarial and deep fuzzy neural network},
  author={Ahmadieh, Hajar and Gassemi, Farnaz and Moradi, Mohammad Hasan},
  journal={Biomedical Signal Processing and Control},
  volume={87},
  pages={105497},
  year={2024},
  publisher={Elsevier}
}

@inproceedings{bai2024dreamdiffusion,
  title={DreamDiffusion: High-quality EEG-to-image generation with temporal masked signal modeling and CLIP alignment},
  author={Bai, Yunpeng and Wang, Xintao and Cao, Yan-Pei and Ge, Yixiao and Yuan, Chun and Shan, Ying},
  booktitle={European Conference on Computer Vision},
  pages={472--488},
  year={2024},
  organization={Springer}
}

@inproceedings{lopez2025guess,
  title={Guess what i think: Streamlined eeg-to-image generation with latent diffusion models},
  author={Lopez, Eleonora and Sigillo, Luigi and Colonnese, Federica and Panella, Massimo and Comminiello, Danilo},
  booktitle={ICASSP 2025-2025 IEEE International Conference on Acoustics, Speech and Signal Processing (ICASSP)},
  pages={1--5},
  year={2025},
  organization={IEEE}
}

@inproceedings{takagi2023high,
  title={High-resolution image reconstruction with latent diffusion models from human brain activity},
  author={Takagi, Yu and Nishimoto, Shinji},
  booktitle={Proceedings of the IEEE/CVF conference on computer vision and pattern recognition},
  pages={14453--14463},
  year={2023}
}

@article{lu2025unfolding,
  title={Unfolding spatiotemporal representations of 3D visual perception in the human brain},
  author={Lu, Zitong and Golomb, Julie D},
  journal={bioRxiv},
  year={2025}
}

@article{welchman2016human,
  title={The human brain in depth: how we see in 3D},
  author={Welchman, Andrew E},
  journal={Annual review of vision science},
  volume={2},
  number={1},
  pages={345--376},
  year={2016},
  publisher={Annual Reviews}
}

@inproceedings{fu2025brainvis,
  title={Brainvis: Exploring the bridge between brain and visual signals via image reconstruction},
  author={Fu, Honghao and Wang, Hao and Chin, Jing Jih and Shen, Zhiqi},
  booktitle={ICASSP 2025-2025 IEEE International Conference on Acoustics, Speech and Signal Processing (ICASSP)},
  pages={1--5},
  year={2025},
  organization={IEEE}
}

@inproceedings{huo2024neuropictor,
  title={Neuropictor: Refining fmri-to-image reconstruction via multi-individual pretraining and multi-level modulation},
  author={Huo, Jingyang and Wang, Yikai and Wang, Yun and Qian, Xuelin and Li, Chong and Fu, Yanwei and Feng, Jianfeng},
  booktitle={European Conference on Computer Vision},
  pages={56--73},
  year={2024},
  organization={Springer}
}

@inproceedings{singh2024learning,
  title={Learning robust deep visual representations from EEG brain recordings},
  author={Singh, Prajwal and Dalal, Dwip and Vashishtha, Gautam and Miyapuram, Krishna and Raman, Shanmuganathan},
  booktitle={Proceedings of the IEEE/CVF Winter Conference on Applications of Computer Vision},
  pages={7553--7562},
  year={2024}
}

@article{cao2025eeg,
  title={Eeg-clip: A transformer-based framework for eeg-guided image generation},
  author={Cao, Xuhao and Gong, Peiliang and Zhang, Liying and Zhang, Daoqiang},
  journal={Neural Networks},
  pages={108167},
  year={2025},
  publisher={Elsevier}
}

@inproceedings{singh2023eeg2image,
  title={EEG2IMAGE: image reconstruction from EEG brain signals},
  author={Singh, Prajwal and Pandey, Pankaj and Miyapuram, Krishna and Raman, Shanmuganathan},
  booktitle={ICASSP 2023-2023 IEEE International Conference on Acoustics, Speech and Signal Processing (ICASSP)},
  pages={1--5},
  year={2023},
  organization={IEEE}
}

@inproceedings{guo2025neuro,
  title={Neuro-3d: Towards 3d visual decoding from eeg signals},
  author={Guo, Zhanqiang and Wu, Jiamin and Song, Yonghao and Bu, Jiahui and Mai, Weijian and Zheng, Qihao and Ouyang, Wanli and Song, Chunfeng},
  booktitle={Proceedings of the Computer Vision and Pattern Recognition Conference},
  pages={23870--23880},
  year={2025}
}

@inproceedings{masclef2025dual,
  title={Dual-Stream EEG Decoding for 3D Visual Perception},
  author={Masclef, Ninon Liz{\'e} and Demcenko, Taisija and Catanzaro, Antonella and Kosmyna, Nataliya},
  booktitle={NeurIPS 2025 Workshop on Symmetry and Geometry in Neural Representations},
  year={2025}
}

@article{deng2025mind2matter,
  title={Mind2Matter: Creating 3D models from EEG signals},
  author={Deng, Xia and Chen, Shen and Zhou, Jiale and Li, Lei},
  journal={arXiv preprint arXiv:2504.11936},
  year={2025}
}

@inproceedings{spampinato2017deep,
  title={Deep learning human mind for automated visual classification},
  author={Spampinato, Concetto and Palazzo, Simone and Kavasidis, Isaak and Giordano, Daniela and Souly, Nasim and Shah, Mubarak},
  booktitle={Proceedings of the IEEE conference on computer vision and pattern recognition},
  pages={6809--6817},
  year={2017}
}

@article{palazzo2020decoding,
  title={Decoding brain representations by multimodal learning of neural activity and visual features},
  author={Palazzo, Simone and Spampinato, Concetto and Kavasidis, Isaak and Giordano, Daniela and Schmidt, Joseph and Shah, Mubarak},
  journal={IEEE Transactions on Pattern Analysis and Machine Intelligence},
  volume={43},
  number={11},
  pages={3833--3849},
  year={2020},
  publisher={IEEE}
}

@inproceedings{deng2009imagenet,
  title={Imagenet: A large-scale hierarchical image database},
  author={Deng, Jia and Dong, Wei and Socher, Richard and Li, Li-Jia and Li, Kai and Fei-Fei, Li},
  booktitle={2009 IEEE conference on computer vision and pattern recognition},
  pages={248--255},
  year={2009},
  organization={Ieee}
}

@article{grattafiori2024llama,
  title={The llama 3 herd of models},
  author={Grattafiori, Aaron and Dubey, Abhimanyu and Jauhri, Abhinav and Pandey, Abhinav and Kadian, Abhishek and Al-Dahle, Ahmad and Letman, Aiesha and Mathur, Akhil and Schelten, Alan and Vaughan, Alex and others},
  journal={arXiv preprint arXiv:2407.21783},
  year={2024}
}

@misc{patil2022stable,
  author       = {Suraj Patil and Pedro Cuenca and Nathan Lambert and Patrick von Platen},
  title        = {Stable Diffusion with Diffusers},
  howpublished = {\url{https://huggingface.co/blog/stable\_diffusion}},
  year         = {2022},
  note         = {Hugging Face Blog},
}

@inproceedings{xiang2025structured,
  title={Structured 3d latents for scalable and versatile 3d generation},
  author={Xiang, Jianfeng and Lv, Zelong and Xu, Sicheng and Deng, Yu and Wang, Ruicheng and Zhang, Bowen and Chen, Dong and Tong, Xin and Yang, Jiaolong},
  booktitle={Proceedings of the IEEE/CVF conference on computer vision and pattern recognition},
  pages={21469--21480},
  year={2025}
}

@inproceedings{kneeland2025nsd,
  title={Nsd-imagery: A benchmark dataset for extending fmri vision decoding methods to mental imagery},
  author={Kneeland, Reese and Scotti, Paul S and St-Yves, Ghislain and Breedlove, Jesse and Kay, Kendrick and Naselaris, Thomas},
  booktitle={Proceedings of the Computer Vision and Pattern Recognition Conference},
  pages={28852--28862},
  year={2025}
}

@inproceedings{li2025brain,
  title={Brain-inspired spiking neural networks for energy-efficient object detection},
  author={Li, Ziqi and Gao, Tao and An, Yisheng and Chen, Ting and Zhang, Jing and Wen, Yuanbo and Liu, Mengkun and Zhang, Qianxi},
  booktitle={Proceedings of the Computer Vision and Pattern Recognition Conference},
  pages={3552--3562},
  year={2025}
}

@inproceedings{shah2025sam4em,
  title={SAM4EM: Efficient memory-based two stage prompt-free segment anything model adapter for complex 3D neuroscience electron microscopy stacks},
  author={Shah, Uzair and Agus, Marco and Boges, Daniya and Chiappini, Vanessa and Alzubaidi, Mahmood and Schneider, Jens and Hadwiger, Markus and Magistretti, Pierre J and Househ, Mowafa and Cal{\`\i}, Corrado},
  booktitle={Proceedings of the IEEE/CVF Conference on Computer Vision and Pattern Recognition},
  pages={4717--4726},
  year={2025}
}

@article{schmors2025trace,
  title={TRACE: Contrastive learning for multi-trial time-series data in neuroscience},
  author={Schmors, Lisa and Gonschorek, Dominic and B{\"o}hm, Jan Niklas and Qiu, Yongrong and Zhou, Na and Kobak, Dmitry and Tolias, Andreas and Sinz, Fabian and Reimer, Jacob and Franke, Katrin and others},
  journal={arXiv preprint arXiv:2506.04906},
  year={2025}
}

@article{wang2025zebra,
  title={ZEBRA: Towards Zero-Shot Cross-Subject Generalization for Universal Brain Visual Decoding},
  author={Wang, Haonan and Lu, Jingyu and Li, Hongrui and Li, Xiaomeng},
  journal={arXiv preprint arXiv:2510.27128},
  year={2025}
}

@article{go2025vist3a,
  title={VIST3A: Text-to-3D by Stitching a Multi-view Reconstruction Network to a Video Generator},
  author={Go, Hyojun and Narnhofer, Dominik and Bhat, Goutam and Truong, Prune and Tombari, Federico and Schindler, Konrad},
  journal={arXiv preprint arXiv:2510.13454},
  year={2025}
}

@article{poole2022dreamfusion,
  title={Dreamfusion: Text-to-3d using 2d diffusion},
  author={Poole, Ben and Jain, Ajay and Barron, Jonathan T and Mildenhall, Ben},
  journal={arXiv preprint arXiv:2209.14988},
  year={2022}
}

@inproceedings{liu2024one,
  title={One-2-3-45++: Fast single image to 3d objects with consistent multi-view generation and 3d diffusion},
  author={Liu, Minghua and Shi, Ruoxi and Chen, Linghao and Zhang, Zhuoyang and Xu, Chao and Wei, Xinyue and Chen, Hansheng and Zeng, Chong and Gu, Jiayuan and Su, Hao},
  booktitle={Proceedings of the IEEE/CVF conference on computer vision and pattern recognition},
  pages={10072--10083},
  year={2024}
}

@inproceedings{xu2024grm,
  title={Grm: Large gaussian reconstruction model for efficient 3d reconstruction and generation},
  author={Xu, Yinghao and Shi, Zifan and Yifan, Wang and Chen, Hansheng and Yang, Ceyuan and Peng, Sida and Shen, Yujun and Wetzstein, Gordon},
  booktitle={European Conference on Computer Vision},
  pages={1--20},
  year={2024},
  organization={Springer}
}

@inproceedings{chen2023seeing,
  title={Seeing beyond the brain: Conditional diffusion model with sparse masked modeling for vision decoding},
  author={Chen, Zijiao and Qing, Jiaxin and Xiang, Tiange and Yue, Wan Lin and Zhou, Juan Helen},
  booktitle={Proceedings of the IEEE/CVF conference on computer vision and pattern recognition},
  pages={22710--22720},
  year={2023}
}

@article{guo2025survey,
  title={A survey of fmri to image reconstruction},
  author={Guo, Weiyu and Sun, Guoying and He, JianXiang and Shao, Tong and Wang, Shaoguang and Chen, Ziyang and Hong, Meisheng and Sun, Ying and Xiong, Hui},
  journal={arXiv preprint arXiv:2502.16861},
  year={2025}
}

@article{bobrov2011brain,
  title={Brain-computer interface based on generation of visual images},
  author={Bobrov, Pavel and Frolov, Alexander and Cantor, Charles and Fedulova, Irina and Bakhnyan, Mikhail and Zhavoronkov, Alexander},
  journal={PloS one},
  volume={6},
  number={6},
  pages={e20674},
  year={2011},
  publisher={Public Library of Science San Francisco, USA}
}

@article{xiang2025electroencephalography,
  title={Electroencephalography-driven three-dimensional object decoding with multi-view perception diffusion},
  author={Xiang, Xin and Zhou, Wenhui and Dai, Guojun},
  journal={Engineering Applications of Artificial Intelligence},
  volume={156},
  pages={111180},
  year={2025},
  publisher={Elsevier}
}
\end{document}